\pdfoutput=1
\documentclass[letterpaper,journal]{IEEEtran}

\usepackage{amsmath}
\usepackage{amssymb}
\usepackage{bm}
\usepackage{booktabs}
\usepackage{cite}
\usepackage{multirow}
\usepackage{graphicx}
\usepackage{color, colortbl}
\usepackage{xcolor}
\usepackage[hidelinks]{hyperref}
\usepackage{cleveref}

\crefname{figure}{Fig.}{Figs.}
\Crefname{figure}{Fig.}{Figs.}

\definecolor{tablered}{RGB}{255, 150, 150}
\definecolor{tableorange}{RGB}{255, 178, 102}
\definecolor{tableyellow}{RGB}{255, 255, 153}
\newcommand{\best}{\cellcolor{tablered}}
\newcommand{\sbest}{\cellcolor{tableorange}}
\newcommand{\tbest}{\cellcolor{tableyellow}}
\newcommand{\textbest}{\colorbox{tablered}{\quad}}
\newcommand{\textsbest}{\colorbox{tableorange}{\quad}}
\newcommand{\texttbest}{\colorbox{tableyellow}{\quad}}

\begin{document}

\title{Uncertainty-Aware Normal-Guided Gaussian Splatting for Surface Reconstruction from Sparse Image Sequences}

\author{Zhen~Tan$^{*}$,
Qiang~Gao$^{*}$,
Xieyuanli~Chen,
Jinpu~Zhang,
Lei~Feng,
Jiaxiong~Liu,
Xiangkai~Lian,
and~Dewen~Hu$^{\dagger}$%
\thanks{$^{*}$Zhen Tan and Qiang Gao contributed equally to this work. $^{\dagger}$Corresponding author: Dewen Hu (dwhu@nudt.edu.cn).}%
\thanks{Zhen Tan, Qiang Gao, Xieyuanli Chen, Jinpu Zhang, Lei Feng, Jiaxiong Liu, Xiangkai Lian, and Dewen Hu are with the College of Intelligence Science and Technology, National University of Defense Technology, Changsha, Hunan 410073, China. E-mail: tz\_paul1996@163.com, qiang\_gao@nudt.edu.cn, xieyuanli.chen@nudt.edu.cn, zhangjinpu@nudt.edu.cn, fenglei20@nudt.edu.cn, liujiaxiong21@nudt.edu.cn, lianxiangkai@nudt.edu.cn, dwhu@nudt.edu.cn.}}

\maketitle

\begin{abstract}
3D Gaussian Splatting (3DGS) has achieved impressive rendering performance in novel view synthesis. 
However, its efficacy diminishes considerably in sparse image sequences, where inherent data sparsity amplifies geometric uncertainty during optimization.
This often leads to convergence at suboptimal local minima, resulting in noticeable structural artifacts in the reconstructed scenes.
To mitigate these issues, we propose Uncertainty-aware Normal-Guided Gaussian Splatting (UNG-GS), a novel framework featuring an explicit Spatial Uncertainty Field (SUF) to quantify geometric uncertainty within the 3DGS pipeline.
UNG-GS enables high-fidelity rendering and achieves high-precision reconstruction without relying on priors.
Specifically, we first integrate Gaussian-based probabilistic modeling into the training of 3DGS to optimize the SUF, providing the model with adaptive error tolerance.
An uncertainty-aware depth rendering strategy is then employed to weight depth contributions based on the SUF, effectively reducing noise while preserving fine details.
Furthermore, an uncertainty-guided normal refinement method adjusts the influence of neighboring depth values in normal estimation, promoting robust results. 
Extensive experiments demonstrate that UNG-GS significantly outperforms leading approaches in sparse-sequence reconstruction.
\end{abstract}


\section{Introduction}
\label{sec:intro}
\IEEEPARstart{H}{igh-quality} surface reconstruction and novel view synthesis are fundamental tasks in computer vision and robotics, with wide-ranging applications in SLAM, 3D reconstruction, AR/VR, and 3D content generation. Recent advancements in Neural Radiance Fields (NeRF)~\cite{mildenhall2020nerf, yang2022recursive, wang2025improving} and 3D Gaussian Splatting (3DGS)~\cite{kerbl3Dgaussians, yu2024mip, li2025mpgs, fan2025fov} have emerged as dominant approaches to these tasks. 

NeRF achieves impressive rendering fidelity through implicit scene representations. 
However, its reliance on ray sampling and volumetric rendering leads to substantial training times, often requiring tens of hours~\cite{wang2021neus, li2023neuralangelo, Fu2022geoneus}.
While 3D Gaussian Splatting (3DGS)~\cite{kerbl3Dgaussians} enables real-time, high-fidelity novel view synthesis using explicit Gaussian point clouds and differentiable rasterization, it still faces limitations in geometric accuracy.
This stems from its heavy reliance on photometric loss, which optimizes uncertainty across views but does not guarantee geometric correctness. It often results in Gaussian distributions drifting away from the actual surface, particularly near object boundaries and thin structures.

To address this, recent methods have introduced geometry-aware enhancements to the 3DGS pipeline.
For example, SuGaR~\cite{guedon2023sugar} imposes geometric regularization and DN-Splatter~\cite{turkulainen2024dnsplatter} leverages foundation models for prior-guided reconstruction.
2DGS~\cite{Huang2DGS2024} represents scenes using 2D Gaussian disks and employs surface normal regularization to constrain geometry. 
PGSR~\cite{chen2024pgsr} further incorporates multi-view photometric regularization into the 3DGS optimization framework.
While these approaches improve reconstruction fidelity, they do not explicitly model geometric uncertainty—a critical omission in sparse-sequence scenarios, where accumulated noise from limited viewpoints, occlusions, and complex spatial layouts severely degrades reconstruction quality. Sparse-sequence settings introduce unique challenges such as depth ambiguities and fractured surfaces due to insufficient parallax and edge-region uncertainty.
Without a mechanism to quantify and mitigate such uncertainty, existing methods may either overfit noisy cues or fail to refine uncertain regions.

\begin{figure}[htbp]
    \centering
    \includegraphics[width=\linewidth]{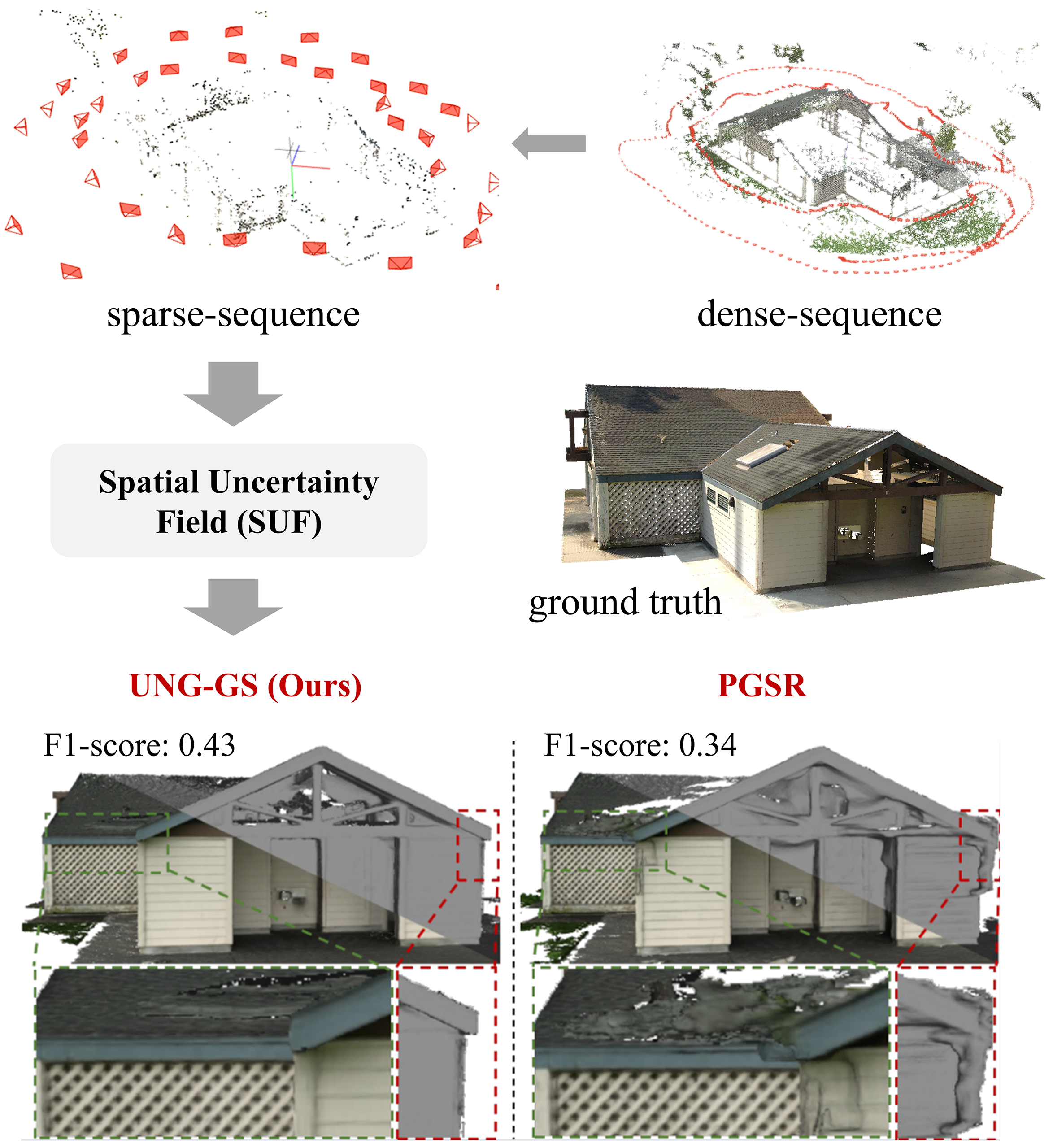}
    \caption{
    Our UNG-GS method explicitly constructs a Spatial Uncertainty Field (SUF) by leveraging normal error during training.
    This design significantly improves surface reconstruction quality in regions with high geometric inconsistency, particularly under sparse-sequence conditions.
    Compared to the state-of-the-art PGSR~\cite{chen2024pgsr}, UNG-GS yields more consistent and smoother reconstructions, especially around object boundaries, effectively mitigating fragmentation artifacts and improving structural coherence.
    }
    \label{fig:teaser}
\end{figure}

In this work, we propose Uncertainty-aware Normal-Guided Gaussian Splatting (UNG-GS), a novel framework to introduce an explicit Spatial Uncertainty Field (SUF) that quantifies geometric uncertainty in 3DGS. As shown in~\Cref{fig:teaser}, this enables high-fidelity reconstruction and rendering without relying on foundation models or additional priors. 
Specifically, we embed probabilistic modeling into the online training framework of 3DGS, equipping the model with adaptive error tolerance through a Gaussian negative log-likelihood loss. 
To suppress noise in uncertain regions, we design an uncertainty-aware rendering strategy that dynamically weights depth rendering based on the SUF. 
Additionally, an uncertainty-guided adaptive normal refinement strategy refines normal estimation by adaptively weighting neighboring depth gradients. This promotes geometric consistency in low-uncertainty areas and reduces sensitivity to the local planar assumption in high-uncertainty regions like edges.

Our key contributions are summarized as follows:
\begin{itemize}
    \item We introduce a novel 3D Gaussian Splatting framework, UNG-GS, incorporating an explicit SUF to achieve high-precision reconstruction and high-fidelity rendering in its default setting without reliance on foundation models or priors;
    \item We propose an uncertainty-aware rendering strategy that dynamically weights depth rendering based on the SUF, effectively suppressing noise in uncertain regions;
    \item We design an uncertainty-guided adaptive normal refinement strategy that adaptively weights gradient calculations for robust normal estimation;
    \item Extensive experiments on DTU and TnT datasets demonstrate that UNG-GS exhibits superior performance over prior works in sparse-sequence reconstruction.
\end{itemize}

\section{Related Work}
\label{sec:related}

\subsection{Neural Radiance Fields}
NeRF~\cite{mildenhall2020nerf} has revolutionized the field of novel view synthesis (NVS) by representing 3D scenes as continuous volumetric functions. NeRF employs a multi-layer perceptron (MLP) to encode both geometry and view-dependent appearance, achieving photorealistic rendering through volume rendering. Subsequent works have extended NeRF in various directions, addressing its limitations in rendering efficiency~\cite{li2023nerfacc} and geometric accuracy~\cite{li2023neuralangelo, wang2021neus, Fu2022geoneus}. 
For instance, Mip-NeRF~\cite{barron2021mip} introduced anti-aliasing techniques to handle multi-scale representations, while Instant-NGP~\cite{muller2022instant} significantly accelerated training through hash-based feature grids. 
Despite these advancements, NeRF-based methods~\cite{li2023neuralangelo, Fu2022geoneus} often struggle with long training times and implicit geometry representations, which limit their ability to produce high-fidelity surface reconstructions. 
NeuS~\cite{wang2021neus} and SparseNeuS~\cite{long2022sparseneus} integrate signed distance functions (SDFs) into the NeRF framework to better extract surfaces, with the latter specifically addressing sparse-view scenarios.
UFORecon~\cite{na2024uforecon} further improves sparse-view reconstruction by incorporating a cross-view matching transformer.
NeRF-W~\cite{martin2021nerfw} and NeRF-on-the-go~\cite{ren2024nerfongo} explore uncertainty modeling within NeRF pipelines to improve robustness in dynamic or uncontrolled environments.
Recent works~\cite{liao2024uncertainty-multi, zhang2023body, goli2024bayes, shen2024estimating3duncertainty} also explore uncertainty modeling in 3D representations.

However, these approaches rely on implicit representations and primarily model uncertainty in appearance or latent space, lacking spatially explicit uncertainty fields for geometry.
In contrast, our approach, UNG-GS, explicitly models geometric uncertainty and leverages a spatial uncertainty field to achieve high-fidelity surface reconstruction under sparse-input settings.

\subsection{3D Gaussian Splatting}
3DGS~\cite{kerbl3Dgaussians} has emerged as a powerful alternative to NeRF, offering real-time rendering and efficient scene representation through explicit 3D Gaussian primitives. 
Unlike NeRF, which relies on volumetric rendering, 3DGS utilizes a differentiable splatting pipeline to project 3D Gaussians onto the image plane, achieving high-quality novel view synthesis with significantly faster training times. 
However, 3DGS faces challenges in accurately representing surfaces due to the volumetric nature of its Gaussian primitives, which can lead to inconsistent geometry across different viewpoints.

Several works have attempted to address these limitations. 
For instance, SuGaR~\cite{guedon2023sugar} introduces additional regularization terms to align Gaussians with surfaces, while DN-Splatter~\cite{turkulainen2024dnsplatter} leverages foundation models to improve reconstruction quality. 
2DGS~\cite{Huang2DGS2024} represents scenes using 2D Gaussian disks and employs surface normal regularization for better surface alignment.
GOF~\cite{Yu2024GOF} extracts the surface by defining an occupancy field exported from the reconstructed 3DGS. 
RaDe-GS~\cite{zhang2024rade} introduces a novel rasterization method for rendering depth and normal maps to improve geometric representation.
PGSR~\cite{chen2024pgsr} designs an unbiased depth estimation rendering method that enables more accurate depth estimation and improves the accuracy of geometric reconstruction by introducing multi-view constraints.
GSDF~\cite{yu2024gsdf} incorporates SDF into the 3DGS optimization to better represent surfaces. MVSplat~\cite{chen2024mvsplat} addresses sparse-view reconstruction using splatting-based architectures, relying on additional supervision or multi-stereo priors.

Despite these improvements, existing methods often fail to explicitly model geometric uncertainty, leading to degraded performance in sparse-sequence scenarios. 
To this end, our proposed UNG-GS framework introduces a Spatial Uncertainty Field (SUF) that explicitly quantifies local confidence in Gaussian geometry.
This uncertainty-aware mechanism enables robust surface reconstruction, suppresses noisy gradients, and provides effective regularization even under limited-view conditions.
\section{Method}
\label{sec:method}
\begin{figure*}[t]
    \centering
    \includegraphics[width=\linewidth]{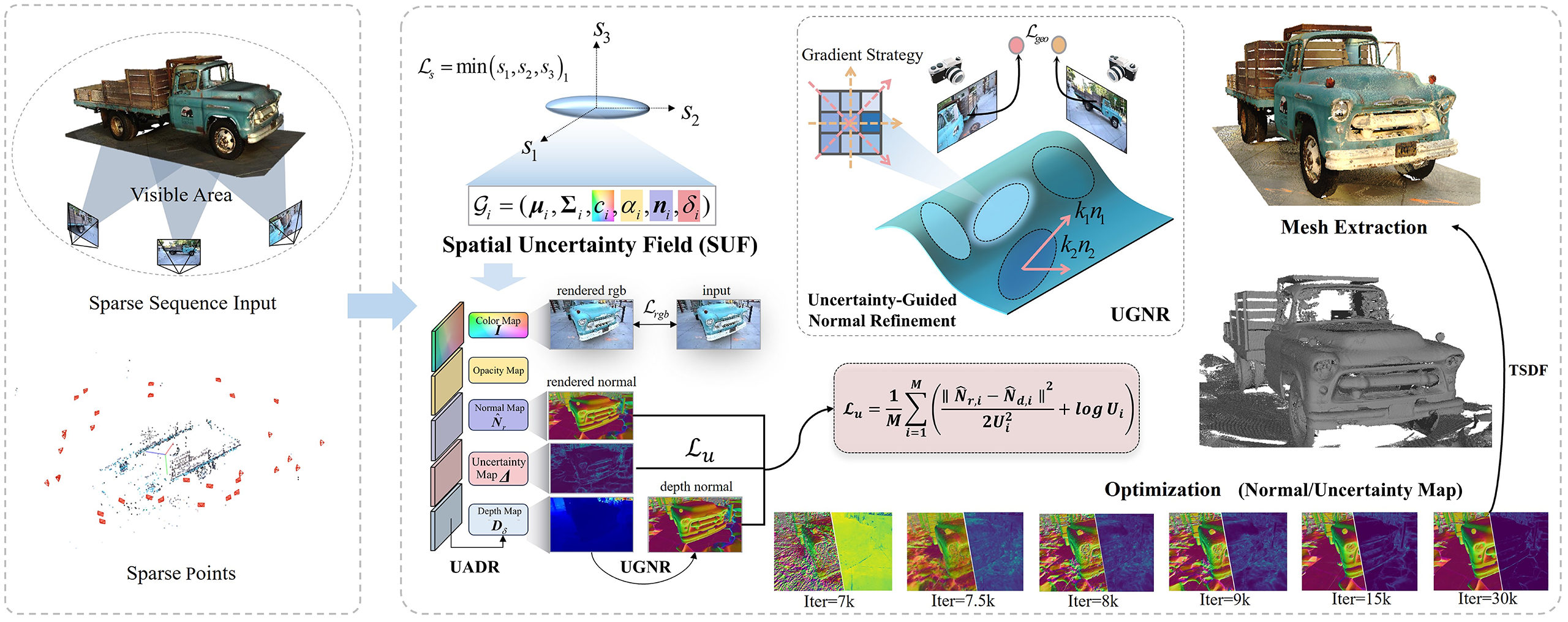}
    \caption{
    \textbf{Overview of UNG-GS framework.}
    Given sparse images, each Gaussian is augmented with an uncertainty score to form a Spatial Uncertainty Field (SUF).
    (a) \textbf{SUF Optimization:} The discrepancy between $\boldsymbol{\hat{N}}_r$ and $\boldsymbol{\hat{N}}_d$ supervises $\mathcal{L}_u$ to learn geometric uncertainty.
    (b) \textbf{Uncertainty-Aware Optimization:} SUF guides depth rendering (UADR) and normal refinement (UGNR), producing refined geometry and final mesh via TSDF.
    }
    \label{fig:pipeline}
\end{figure*}

Our UNG-GS framework, as illustrated in~\Cref{fig:pipeline}, introduces an explicit Spatial Uncertainty Field (SUF) to enhance 3D Gaussian Splatting (3DGS) for robust surface reconstruction, particularly from sparse image sequences.
We first describe a foundation with planar-based 3D Gaussian Splatting (Section \ref{sec:pre}), rendering color, depth, and normals from 3D Gaussians.
To explicitly model geometric uncertainty, we introduce SUF (Section \ref{sec:suf}), which assigns an uncertainty value to each Gaussian.
Next, we leverage SUF in our Uncertainty-Aware Geometric Reconstruction module (Section \ref{sec:geo-recon}), which includes Uncertainty-Aware Depth Rendering (UADR) and Uncertainty-Guided Normal Refinement (UGNR) to improve depth and normal estimation.
Finally, the framework is trained end-to-end using uncertainty-aware loss functions (Section \ref{sec:training}) to optimize rendering quality, geometric accuracy, and uncertainty estimation.

\subsection{Preliminaries}
\label{sec:pre}

\subsubsection{Planar-based 3D Gaussian Splatting}
\label{sec:planar-based-3DGS}
\begin{figure*}[t]
    \centering
    \includegraphics[width=0.95\linewidth]{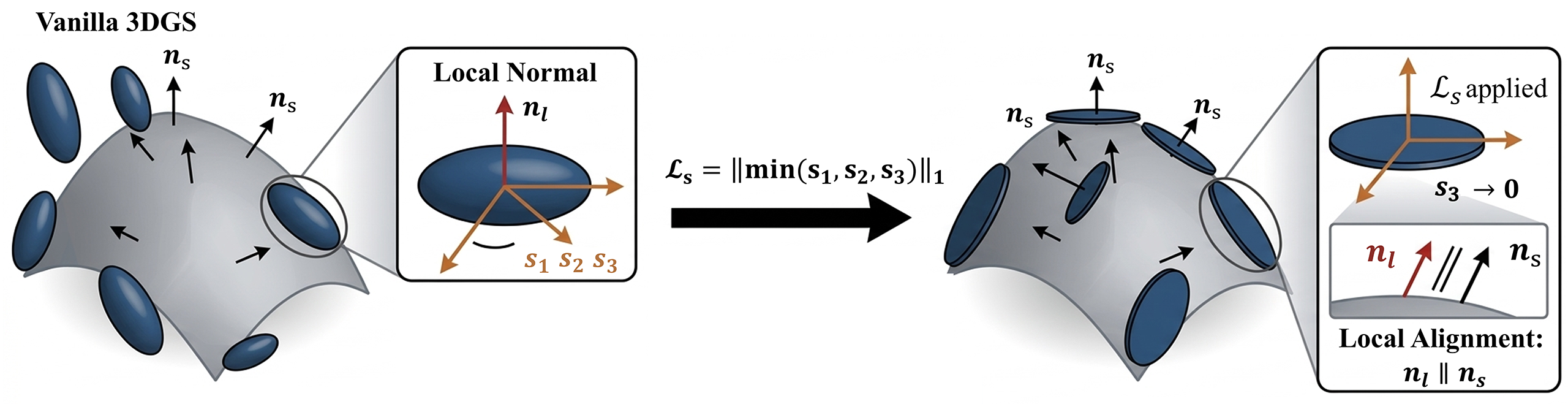}
    \caption{\textbf{Illustration of planar-based 3D Gaussian splatting.} 
    Minimizing the smallest Gaussian scale enforces flattening along the surface normal, yielding a locally planar representation. 
    This aligns the local normal $n_l$ with the surface normal $n_s$, improving geometric accuracy and normal consistency.}
    \label{fig:planar_based_3dgs}
\end{figure*}

3DGS represents a scene as a set of 3D Gaussians $\mathcal{G}_i$, each defined by its center $\boldsymbol{\mu}_i \in \mathbb{R}^3$, covariance matrix $\boldsymbol{\Sigma}_i \in \mathbb{R}^{3 \times 3}$, opacity $\alpha_i$, and color $\boldsymbol{c}_i$. The covariance matrix is factorized into a scaling matrix $\boldsymbol{S}_i$ and a rotation matrix $\boldsymbol{R}_i$ as $\boldsymbol{\Sigma}_i = \boldsymbol{R}_i \boldsymbol{S}_i \boldsymbol{S}_i^T \boldsymbol{R}_i^T$. The Gaussian is then expressed as:
\begin{equation}
\mathcal{G}_i\left(\boldsymbol{x} \mid \boldsymbol{\mu}_i, \boldsymbol{\Sigma}_{\boldsymbol{i}}\right)=e^{-\frac{1}{2}\left(\boldsymbol{x}-\boldsymbol{\mu}_i\right)^{\top} \boldsymbol{\Sigma}_i^{-1}\left(\boldsymbol{x}-\boldsymbol{\mu}_i\right)}.
\end{equation}
Each Gaussian is projected into the camera coordinate system via:
\begin{equation}
\boldsymbol{\hat{\mu}}_i=\boldsymbol{K} \boldsymbol{W}\left[\boldsymbol{\mu}_i, 1\right]^{\top}, \quad \boldsymbol{\hat{\Sigma}}_i=\boldsymbol{J} \boldsymbol{W} \boldsymbol{\Sigma}_i \boldsymbol{W}^{\top} \boldsymbol{J}^{\top}, 
\end{equation}
where $\boldsymbol{W}$, $\boldsymbol{K}$, and $\boldsymbol{J}$ denote the world-to-camera transformation, the camera intrinsic matrix, and the Jacobian of the local affine approximation of the projection.

To encourage planar alignment and improve geometric accuracy, we adopt a planar-based Gaussian representation, inspired by~\cite{chen2024pgsr, chen2023neusg}. We minimize the smallest singular value of the scaling matrix $\boldsymbol{S}_i$ using a regularization term:
\begin{equation}
\mathcal{L}_s=\left\|\min \left(s_1, s_2, s_3\right)\right\|_1,
\end{equation}
where $s_1$, $s_2$, and $s_3$ are the singular values of the scaling matrix $\boldsymbol{S}_i$. As illustrated in~\Cref{fig:planar_based_3dgs}, this regularization encourages Gaussians to flatten along the surface-normal direction, ensuring their centers align closely with the underlying geometry.

\subsubsection{Color, Depth, and Normal Rendering}
\label{sec:rendering}
Following typical 3DGS pipelines~\cite{kerbl3Dgaussians}, the rendered color $\boldsymbol{I}$ is obtained via $\alpha$-compositing:
\begin{equation}
\boldsymbol{I}=\sum_{i \in N} T_i \hat{\alpha}_i \boldsymbol{c}_i, \quad T_i=\prod_{j=1}^{i-1}\left(1-\hat{\alpha}_j\right),
\end{equation}
where $N$ denotes Gaussians sorted by depth, $T_i$ represents the accumulated transmittance, and $\hat{\alpha}$ is calculated by evaluating $\mathcal{G}_i\left(\boldsymbol{x} \mid \boldsymbol{\hat{\mu}}_i, \boldsymbol{\hat{\Sigma}}_{\boldsymbol{i}}\right)$ multiplied by a learnable opacity.
Similarly, the rendered normal map $\boldsymbol{\tilde{N}}_r$ is generated by blending the normals of the Gaussian planes:
\begin{equation}
\boldsymbol{\tilde{N}}_r=\sum_{i \in N} T_i \hat{\alpha}_i \boldsymbol{R}_c^T \boldsymbol{n}_i, 
\end{equation}
and its normalized counterpart is $\boldsymbol{\hat{N}}_r = \boldsymbol{\tilde{N}}_r / \|\boldsymbol{\tilde{N}}_r\|$,
where $\boldsymbol{R}_c$ denotes the rotation from the camera to the global world coordinate system. To obtain unbiased depth~\cite{chen2024pgsr, shao2024nddepth}, the distance from the plane to the camera center can be represented as $d_i=\left(\boldsymbol{R}_c^T\left(\boldsymbol{\mu}_i-\boldsymbol{O}_c\right)\right) \boldsymbol{R}_c^T \boldsymbol{n}_i^T$, where $\boldsymbol{O}_c$ is the camera center in the world. Then, the distance map $\boldsymbol{Dist}$ and unbiased depth $\boldsymbol{D}$ can be computed as:
\begin{equation}
\boldsymbol{D}=\frac{\boldsymbol{Dist}}{\boldsymbol{\tilde{N}}_r \boldsymbol{K}^{-1} \tilde{\boldsymbol{p}}} , \quad \boldsymbol{Dist}=\sum_{i \in N} T_i \hat{\alpha}_i d_i,
\end{equation}
where $\tilde{\boldsymbol{p}}$ denotes the homogeneous coordinates of the pixel.

\subsection{Spatial Uncertainty Field (SUF)}
\label{sec:suf}
\begin{figure}[t]
    \centering
    \includegraphics[width=\linewidth]{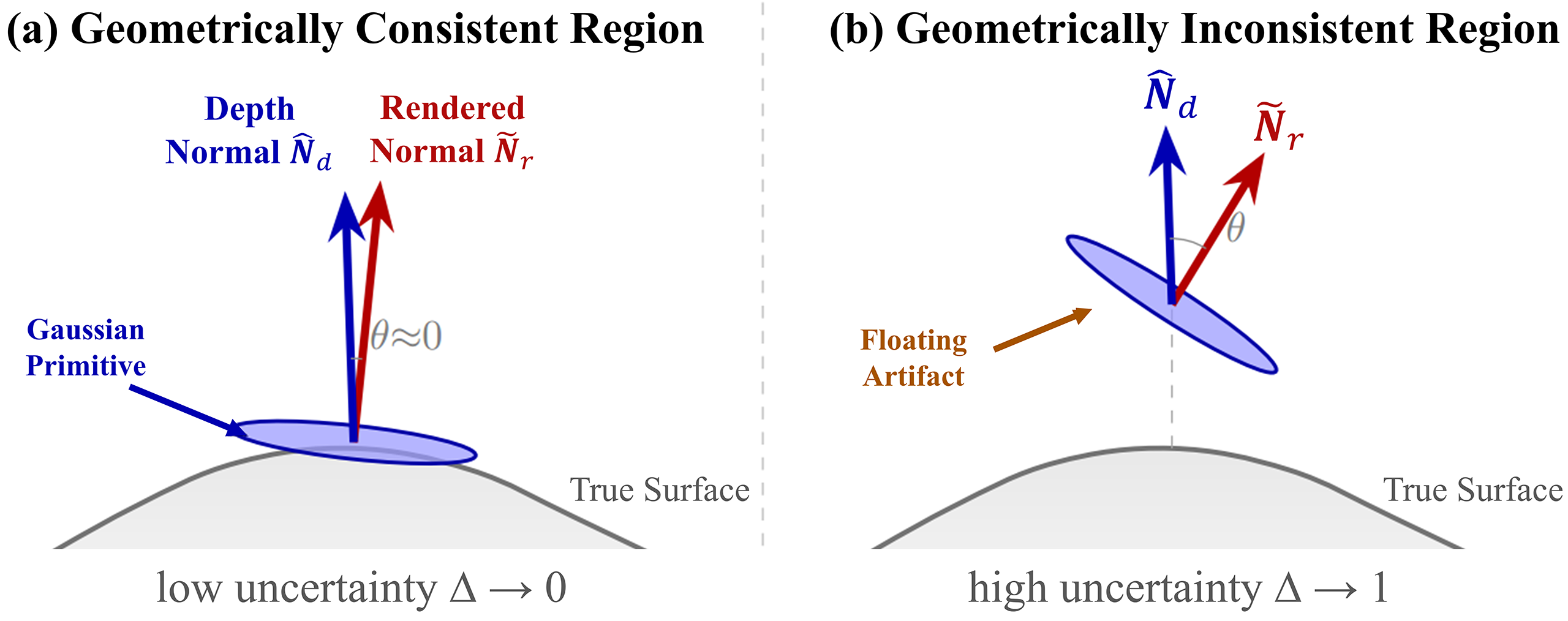}
    \caption{\textbf{Illustration of the spatial uncertainty field (SUF).} 
    (a) In consistent regions, $\hat{\mathbf{N}}_r$ aligns with $\hat{\mathbf{N}}_d$, yielding low uncertainty ($\Delta \rightarrow 0$). 
    (b) In inconsistent regions, their deviation increases ($\Delta \rightarrow 1$), indicating high uncertainty.}
    \label{fig:suf}
\end{figure}

With limited input images, such as sparse sequences from a moving camera or restricted coverage scenarios, uncertainty arises from limited viewpoints, occlusions, and noisy depth estimates, leading to ambiguities in localization and surface reconstruction. To explicitly capture this uncertainty, we assign each 3D Gaussian an additional scalar attribute, $u_i \in [0,1]$, representing its geometric uncertainty. Intuitively, $u_i$ measures how reliably the geometry of $\mathcal{G}_i$ is supported by the available views, and it is optimized end-to-end from the discrepancy between the normalized rendered normal and the depth-based normal after rasterization, rather than estimated from an explicit posterior distribution. A value of $u_i = 0$ indicates perfect confidence in the Gaussian's geometry, while $u_i = 1$ signifies maximum uncertainty. As illustrated in~\Cref{fig:suf}, regions with good agreement between the rendered normal and the depth-based normal correspond to low uncertainty, whereas large normal discrepancies indicate geometrically inconsistent regions with high uncertainty. Similar to color and depth, we render the uncertainty map $\boldsymbol{U}$ by alpha-compositing the uncertainty values:
\begin{equation}
\boldsymbol{U}=\sum_{i \in N} T_i \hat{\alpha}_i u_i. 
\end{equation}

To effectively train and optimize SUF, we propose an uncertainty normal loss inspired by the Gaussian Negative Log-Likelihood (NLL) framework:
\begin{equation}
\mathcal{L}_u=\frac{1}{M} \sum_{i=1}^{M}\left(\frac{\left\|\boldsymbol{\hat{N}}_{r,i}-\boldsymbol{\hat{N}}_{d,i}\right\|^2}{2\,U_i^2}+\log U_i\right),
\label{eq:uncertainty_loss}
\end{equation}
where $M$ denotes the total number of valid pixels whose accumulated opacity exceeds a threshold, $\boldsymbol{\hat{N}}_{r,i}$ and $\boldsymbol{\hat{N}}_{d,i}$ are the normalized rendered normal and depth-based normal at pixel $i$, and $U_i$ is the rendered uncertainty scale at pixel $i$ obtained by alpha-compositing the per-Gaussian uncertainties that contribute to pixel $i$.
In practice, $U_i$ is clamped to $[\epsilon,\,1]$ with $\epsilon=10^{-6}$ for numerical stability.
We model the uncertainty loss with an NLL-inspired objective~\cite{nix1994estimating, barron2019general}, using the mismatch between $\boldsymbol{\hat{N}}_{r,i}$ and $\boldsymbol{\hat{N}}_{d,i}$ as the training signal.
Here, $U_i$ is a learned uncertainty scale similar to a heteroscedastic standard-deviation term, but it remains bounded through the Gaussian attribute $u_i \in [0,1]$.
We therefore interpret $U_i$ as a bounded measure of geometric confidence rather than an unconstrained variance.
The first term in~\Cref{eq:uncertainty_loss} penalizes under-estimated uncertainty when the normal error is large, whereas the $\log(U_i)$ term prevents the model from assigning unnecessarily large uncertainty values.

\subsection{Uncertainty-Aware Geometric Reconstruction}
\label{sec:geo-recon}
\subsubsection{Uncertainty-Aware Depth Rendering (UADR)}
To enhance depth rendering with SUF, we propose an uncertainty-aware depth rendering strategy. It adjusts each Gaussian's contribution to the final depth through a Confidence Modulation Function (CMF), which maps the per-Gaussian uncertainty to a confidence weight:
\begin{equation}
w_i=-0.5 u_i^2+1 .
\end{equation}
This quadratic CMF changes the depth weight smoothly as the uncertainty increases.
For a confident Gaussian ($u_i = 0$), the weight remains $w_i = 1$.
As $u_i$ becomes larger, the weight decreases with derivative $\frac{\mathrm{d}w_i}{\mathrm{d}u_i} = -u_i$, so uncertain Gaussians are down-weighted progressively rather than abruptly, which improves optimization stability (\Cref{sec:ablation}).
At the same time, even the most uncertain Gaussians still keep a residual weight of $0.5$, so potentially useful information is not discarded entirely.
Relative to the alternative modulation functions used in our experiments, this behavior is more effective at reducing noisy depth contributions while keeping geometric details.

The uncertainty-aware rendered depth is computed as:
\begin{equation}
\boldsymbol{D_u}=\sum_{i \in N} T_i \hat{\alpha}_i d_i w_i.
\end{equation}
This ensures that Gaussians with low uncertainty contribute more significantly to the final depth map. In contrast, those with high uncertainty are suppressed, effectively reducing noise in uncertain regions.

\subsubsection{Uncertainty-Guided Normal Refinement (UGNR)}
To enhance the robustness of depth-based normal estimation, we introduce an uncertainty-guided normal refinement strategy. Because the depth-based normals are derived from $\boldsymbol{D}_u$ via finite differences, uncertainty weighting is inherently propagated into the normal estimates. Specifically, we compute two sets of cross products from neighboring depth gradients:
\begin{equation}
\boldsymbol{n}_1=\nabla_x \boldsymbol{D}_u \times \nabla_y \boldsymbol{D}_u, \quad \boldsymbol{n}_2=\nabla_{d 1} \boldsymbol{D}_u \times \nabla_{d 2} \boldsymbol{D}_u,
\end{equation}
where $\nabla_x$ and $\nabla_y$ are the horizontal and vertical gradients of the depth map, $\nabla_{d 1}$ and $\nabla_{d 2}$ are the gradients along the two diagonal directions. The final refined depth-based normal $\boldsymbol{\hat{N}}_d$ is computed as a weighted average of $\boldsymbol{n_1}$ and $\boldsymbol{n_2}$:
\begin{equation}
\boldsymbol{\hat{N}}_d=\frac{k_1 \boldsymbol{n}_1+k_2 \boldsymbol{n}_2}{\left\|k_1 \boldsymbol{n}_1+k_2 \boldsymbol{n}_2\right\|}, \quad k_j=\sum_{p \in \mathcal{P}_j}\bigl(1-U(p)\bigr),\; j \in \{1,2\},
\end{equation}
where the weight $k_j$ is determined by the uncertainty values of the pixels involved in the gradient calculations, $\mathcal{P}_j$ is the set of pixels used to compute $\boldsymbol{n}_j$, and $U(p)$ denotes the rendered uncertainty scale at pixel $p$. This weighting scheme ensures that regions with lower uncertainty contribute more to the final normal estimation, enhancing robustness in high-uncertainty areas such as edges.

\subsection{Training Functions}
\label{sec:training}
To effectively train UNG-GS, we employ four loss functions: RGB loss, multi-view consistency loss, scale regularization, and uncertainty normal loss.
\subsubsection{RGB Loss}
To account for variations in image brightness caused by changing external lighting conditions over time, we incorporate an exposure model~\cite{martin2021nerfw} into the training procedure. By computing the exposure coefficients $m_i$ and $b_i$, we obtain the $i$-th exposure-compensated image: $\boldsymbol{I}_i^e=\exp \left(m_i\right) \boldsymbol{I}_i+b_i$. 
To ensure structural similarity, the final RGB loss is formulated as:
\begin{equation}
\mathcal{L}_{r g b}=(1-\lambda) \mathcal{L}_1\left({\tilde{\boldsymbol{I}^e}}-\boldsymbol{\hat{I}}\right)+\lambda \mathcal{L}_{S S I M}\left(\boldsymbol{I}-\boldsymbol{\hat{I}}\right),
\end{equation}
where $\tilde{\boldsymbol{I}^e}$ is the rendered image adjusted by the exposure coefficients. If the SSIM difference is less than 0.5, the original rendered image is used directly: $\tilde{\boldsymbol{I}^e} = \boldsymbol{I}$. 
$\boldsymbol{\hat{I}}$ denotes the input image, $\mathcal{L}_1$ denotes the L1 loss constraint, and $\mathcal{L}_{S S I M}$ denotes the SSIM loss. 
$\lambda$ is set to 0.2.

\subsubsection{Multi-View Consistency Loss}
In the case of sparse image sequences, COLMAP~\cite{schoenberger2016colmap} generates only a sparse initial point cloud. When relying solely on the single-view constraint, the optimization may suffer from geometric inconsistencies. To address this problem and enhance the reconstruction quality, we introduce multi-view consistency constraints~\cite{chen2024pgsr} based on planar patches:
\begin{equation}
\tilde{\boldsymbol{p}}_n=\boldsymbol{H}_{r n} \tilde{\boldsymbol{p}}_r, \quad \boldsymbol{H}_{r n}=\boldsymbol{K}_r\left(\boldsymbol{R}_{r n}-\frac{\boldsymbol{T}_{r n} \boldsymbol{n}_r^T}{d_r}\right) \boldsymbol{K}_r^{-1},
\end{equation}
where $\tilde{\boldsymbol{p}}_r$ and $\tilde{\boldsymbol{p}}_n$ are the correspondences, $\boldsymbol{H}_{r n}$ and $\boldsymbol{T}_{r n}$ are the homography and transformation from the reference frame to the neighboring frame, respectively.  

The multi-view consistency loss $\mathcal{L}_{mv}$ consists of two components: the multi-view photometric constraint $\mathcal{L}_{ncc}$ and the multi-view geometric consistency constraint $\mathcal{L}_{geo}$. Photometric consistency is ensured by minimizing NCC~\cite{yoo2009ncc} between the reference and adjacent views:
\begin{equation}
\mathcal{L}_{n c c}=\frac{1}{|\mathbb{V}|} \sum_{\boldsymbol{p}_r \in \mathbb{V}}\left(1-N C C\left(I\left(\boldsymbol{p}_r\right), I\left(\boldsymbol{H}_{n r} \boldsymbol{p}_n\right)\right)\right),
\end{equation}
where $\mathbb{V}$ denotes the valid region. Geometric consistency is achieved by minimizing reprojection errors:
\begin{equation}
\mathcal{L}_{g e o}=\frac{1}{|\mathbb{V}|} \sum_{\boldsymbol{p}_r \in \mathbb{V}}\left\|\tilde{\boldsymbol{p}}_r-\boldsymbol{H}_{n r} \boldsymbol{H}_{r n} \tilde{\boldsymbol{p}}_r\right\|,
\end{equation}
and the combined multi-view consistency loss is:
\begin{equation} 
\mathcal{L}_{mv} = \mathcal{L}_{ncc} + \mathcal{L}_{geo},
\end{equation}

\subsubsection{Total Loss}
In summary, our complete training loss is:
\begin{equation}
\mathcal{L}=\mathcal{L}_{r g b}+ \lambda_1 \mathcal{L}_u + \lambda_2 \mathcal{L}_s+\mathcal{L}_{mv}, 
\end{equation}
where $\lambda_1$ and $\lambda_2$ weigh the importance of the uncertainty and scaling regularization term, respectively.

\section{Experiments}
\label{sec:experiments}

\subsection{Experimental Setup}
\label{sec:exp setup}
\subsubsection{Datasets} To evaluate our method, we conduct experiments on the Tanks \& Temples (TnT)~\cite{knapitsch2017tnt}, DTU~\cite{jensen2014dtu}, and Mip-NeRF360~\cite{barron2022mip360} datasets.
To assess reconstruction performance under the challenging sparse-sequence conditions, we simulate this setting on the TnT and DTU datasets. 
Specifically, we downsample the image sequences, retaining one frame every eight for TnT and every four for DTU, respectively.
Sparse point clouds and camera poses are then generated using COLMAP~\cite{schoenberger2016colmap}. 
For the novel view synthesis, we evaluate rendering quality on the Mip-NeRF360  dataset, using its standard setup as it is a dedicated benchmark for this purpose.

\subsubsection{Implementation}
We design a novel uncertainty-aware differentiable rasterization module, implemented in CUDA, which generalizes and enhances the capabilities of 3DGS and PGSR by explicitly modeling uncertainty in the rendering process. We employ AbsGS's~\cite{ye2024absgs} densification strategy and extract meshes via TSDF Fusion~\cite{curless1996tsdf}. Our training strategy has three stages: (1) only photometric loss and scale regularization before 15,000 iterations; (2) uncertainty normal loss and multi-view constraints to optimize spatial uncertainty field (SUF) and geometry from iteration 15,000, where $\boldsymbol{\hat{N}}_d$ is computed from the standard rendered depth $\boldsymbol{D}$ via finite differences; and (3) UGNR starting at iteration 10,000, after which $\boldsymbol{\hat{N}}_d$ is replaced by the uncertainty-refined normal for enhanced geometric quality. All experiments are conducted on the NVIDIA RTX 4090 GPU, with the uncertainty learning rate set to 0.0025. On the TnT dataset, we use $\lambda_1$ = 0.008, while on the DTU dataset, $\lambda_1$ = 1e-5. The $\lambda_2$ is set to 100. Our training strategy on the DTU dataset involved introducing multi-view and uncertainty losses at iteration 8000. We adopt a similar strategy to~\cite{chen2024pgsr} for constructing the image graph and selecting adjacent frames, ensuring effective spatial relationships between views.

\setlength\tabcolsep{0.5em}
\begin{table*}[t]
\centering
\caption{\textbf{Quantitative comparison on the DTU Dataset~\cite{jensen2014dtu}}. We report the Chamfer Distance (CD) [mm$\downarrow$] and our method achieves the best performance among all GS-based methods. \textbest, \textsbest, \texttbest ~indicate the best, the second best, and the third best respectively.}
\scriptsize
\setlength{\tabcolsep}{2.6pt}
\renewcommand\arraystretch{1}
\begin{tabular}{@{}l|ccccccccccccccc|cc}
\toprule
 & 24 & 37 & 40 & 55 & 63 & 65 & 69 & 83 & 97 & 105 & 106 & 110 & 114 & 118 & 122 & Mean & Time \\ \midrule
2D GS~\cite{Huang2DGS2024} & 1.19& \tbest 1.06& 1.22& 0.64& 1.38& 1.20& 1.01& 1.27& \tbest 1.22& 0.68& 1.09& 1.86& 0.50& \tbest 0.87& 0.58& 1.05& \sbest9.6m\\
GOF~\cite{Yu2024GOF} & 0.81& 1.10& \tbest 0.81& \tbest 0.46& 1.36& 1.11& 0.82& 1.21& 1.41& \tbest 0.61& 1.05& 1.33& 0.49& 0.88& 0.64& 0.94& 1h\\
RaDe-GS~\cite{zhang2024rade} & \tbest 0.74& 1.15& 1.06& \sbest 0.44& \tbest 
0.95& \tbest 1.02& \tbest 0.78& \tbest 1.20& \tbest 1.22& 0.65& \tbest 0.86& \tbest 0.89& \tbest 0.40& 0.92& \tbest 0.57& \tbest 0.86& \best8m\\
PGSR~\cite{chen2024pgsr} & \sbest 0.72& \sbest 0.86& \sbest 0.75& 0.67& \sbest 0.81& \best 0.70& \sbest 0.56& \sbest 1.06& \sbest 1.00& \sbest 0.59& \sbest 0.63& \sbest 0.51& \sbest 0.31& \sbest 0.38& \sbest 0.36& \sbest 0.66& 0.5h\\
\midrule
Ours & \best 0.64&  \best 0.82& \best 0.63& \best 0.43& \best 0.80& \sbest 0.74& \best 0.53& \best 1.05& \best 0.99& \best 0.58& \best 0.54& \best 0.51& \best 0.30& \best 0.35& \best 0.35& \best 0.62& \tbest25m\\
\bottomrule
\end{tabular}
\label{tab:dtu_result}
\end{table*}

\begin{table}[t]
\centering
\caption{\textbf{Quantitative comparison on the TnT Dataset~\cite{knapitsch2017tnt}}. We report the F1-score and average optimization time. All results are evaluated with the official evaluation scripts. Our method achieves the best score among Gaussian Splatting methods.}
\scriptsize
\setlength{\tabcolsep}{3pt}
\renewcommand\arraystretch{1}
\begin{tabular*}{\columnwidth}{@{\extracolsep{\fill}}lccccc@{}}
\toprule
 & 2D GS~\cite{Huang2DGS2024} & GOF~\cite{Yu2024GOF} & RaDe-GS~\cite{zhang2024rade} & PGSR~\cite{chen2024pgsr} & Ours \\ 
\midrule
Barn & \sbest 0.34 &  0.29 & \tbest 0.30 & \sbest 0.34 &  \best 0.43 
\\
Caterpillar &  0.26 &  0.27 & \tbest 0.28 & \sbest 0.31 & \best 0.35 
\\
Courthouse & \tbest 0.18 & \sbest 0.23 & \best 0.24 &  0.08 & \best 0.24 
\\
Ignatius &  0.45 &  0.50 & \tbest 0.55 & \sbest 0.61 & \best 0.63 
\\
Meetingroom &  0.03 & \sbest 0.11 & \tbest 0.10 & \tbest 0.10 & \best 0.12 
\\
Truck & \tbest 0.39 &  0.32 &  0.33 & \sbest 0.52 & \best 0.56 
\\ 
\midrule
Mean &  0.28 &  0.29 & \tbest 0.30 & \sbest 0.33 & \best 0.39 
\\
Time & \best 15m& \textgreater 1h& \sbest 23m&  1.1h& \tbest 40m\\ 
\bottomrule
\end{tabular*}
\label{tab:tnt}
\end{table}

\subsubsection{Evaluations}
We compare our method with SOTA Gaussian Splatting methods, including 2DGS~\cite{Huang2DGS2024}, GOF~\cite{Yu2024GOF}, RaDe-GS~\cite{zhang2024rade}, and PGSR~\cite{chen2024pgsr}, for surface reconstruction and rendering performance.
Given the high computational cost of NeRF-based methods~\cite{mildenhall2020nerf, wang2021neus}, we restrict the main comparison to Gaussian Splatting-based methods and additionally report results under standard dense-sequence settings as an extended robustness analysis.
We evaluate reconstruction quality by following baselines~\cite{Huang2DGS2024, Yu2024GOF, zhang2024rade, chen2024pgsr} using Chamfer Distance (CD) on the DTU dataset and F1-score on the TnT dataset. 
For novel view synthesis, we employ PSNR, SSIM, and LPIPS as evaluation metrics. Our primary evaluation focuses on sparse-sequence surface reconstruction on DTU and TnT. We use Mip-NeRF 360 as a secondary benchmark to verify that the proposed uncertainty-aware design does not compromise rendering quality. Dense-sequence and prior-assisted experiments are included as extended robustness analyses to assess the scope of the method beyond the main sparse-sequence setting.

\subsection{Results on Sparse-Sequence Reconstruction}
\label{sec:results}

\begin{figure*}[t]
    \centering
    \includegraphics[width=\linewidth]{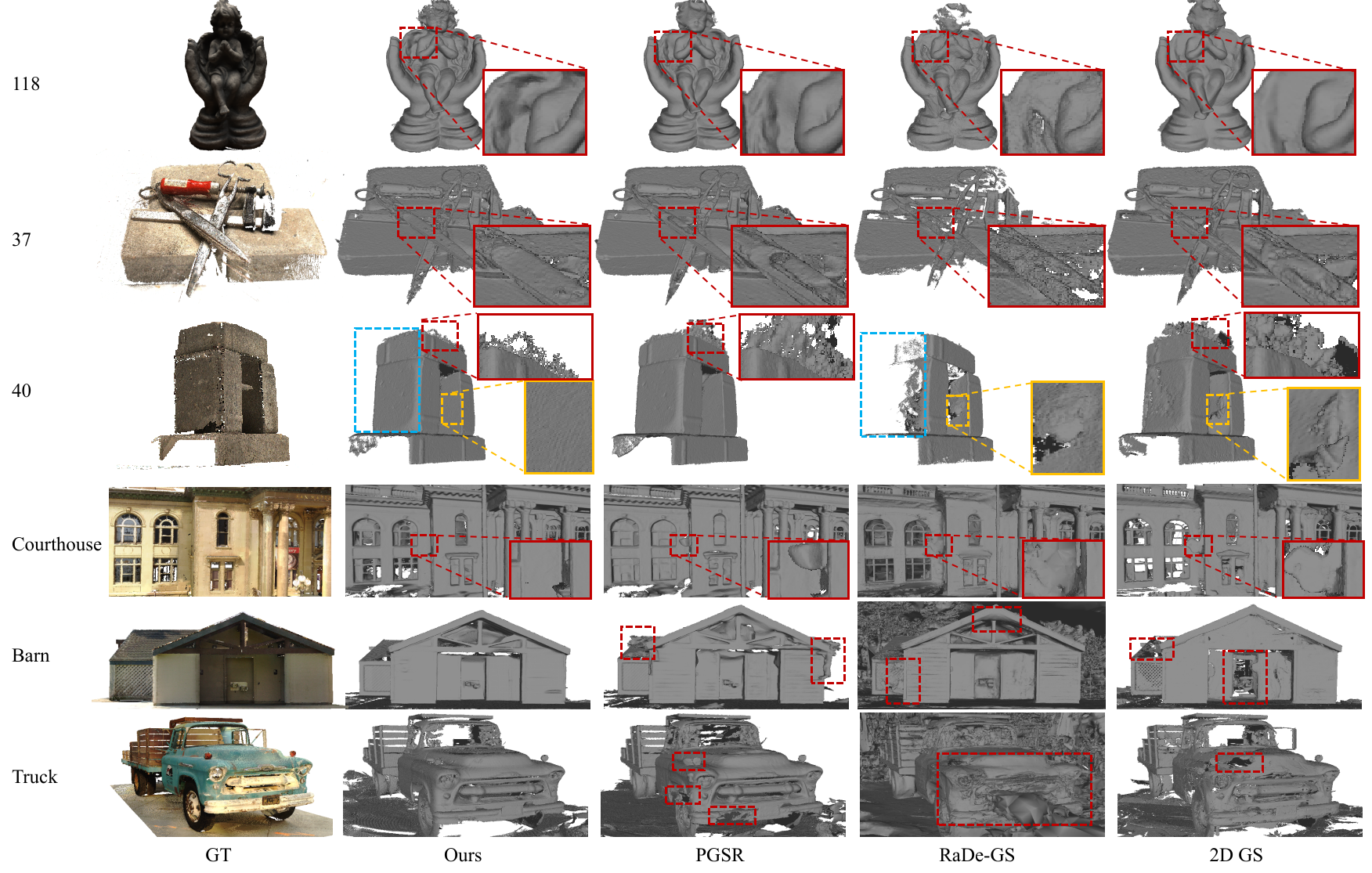}
    \caption{\textbf{The qualitative comparisons of surface reconstruction on the DTU~\cite{jensen2014dtu} and TnT~\cite{knapitsch2017tnt} datasets. }Our method produces smoother and more detailed surfaces under sparse-sequence compared to existing GS-based methods~\cite{chen2024pgsr, zhang2024rade, Huang2DGS2024}.}
    \label{fig:mesh_recon}
\end{figure*}

\subsubsection{Reconstruction Performance}
\label{sec:recon}
We compare our method with existing methods on the DTU and TnT datasets in~\Cref{tab:dtu_result} and~\Cref{tab:tnt}. On the DTU dataset, our method achieves state-of-the-art reconstruction accuracy, as quantified by an average CD of 0.62, while maintaining a relatively efficient optimization time of 25 minutes. Compared to existing approaches, our method improves reconstruction accuracy by 6.1\% to 41.0\%. On the TnT dataset, our method attains state-of-the-art reconstruction quality, yielding an average F1-score of 0.39 with a competitive optimization time of 40 minutes. Compared to other methods, the reconstruction quality improves by 18.2\% to 39.3\%. ~\Cref{fig:mesh_recon} presents a qualitative comparison with existing GS-based methods on DTU and TnT, where our method achieves higher-quality results. Overall, UNG-GS improves sparse-sequence reconstruction accuracy while keeping the optimization time competitive.

\subsection{Rendering Performance}
\label{sec:rendering-perf}
\subsubsection{Quantitative Results on Mip-NeRF 360}
As a secondary validation, we compare our method with existing methods on the Mip-NeRF 360~\cite{barron2022mip360} dataset. The results in~\Cref{tab:mipnerf360} show that our approach also improves rendering quality in the novel view synthesis task, indicating that the proposed uncertainty-aware design does not compromise rendering fidelity while targeting geometric reconstruction.

\subsubsection{Qualitative Results on Mip-NeRF 360}
\begin{figure*}[t]
    \centering
    \includegraphics[width=\linewidth]{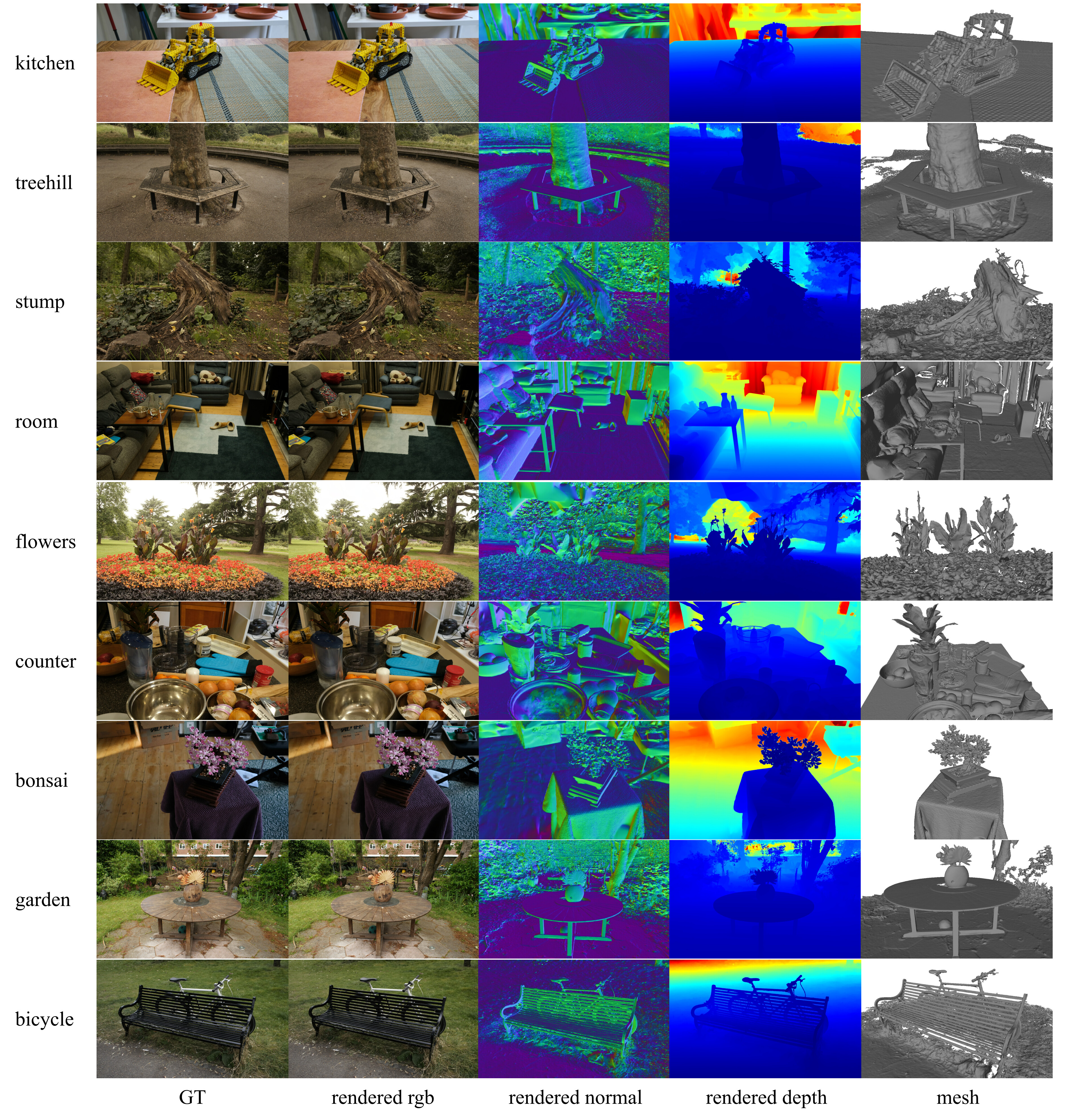}
    \caption{\textbf{Qualitative results on Mip-NeRF 360~\cite{barron2022mip360} dataset.} Our approach can generate high-quality meshes and images including RGB, normal, and depth.}
    \label{fig:mip-nerf}
\end{figure*}

In addition to the quantitative rendering results, we provide qualitative results on the Mip-NeRF 360 dataset. As shown in~\Cref{fig:mip-nerf}, our method produces high-quality RGB renderings, depth maps, normal maps, and meshes. These results further demonstrate that UNG-GS maintains a favorable balance between rendering fidelity and geometric accuracy, yielding visually coherent scene appearance while preserving detailed surface structures.
\begin{table}
\centering
\caption{\textbf{Quantitative results on Mip-NeRF 360~\cite{barron2022mip360} dataset.}}
\scriptsize
\setlength{\tabcolsep}{3pt}
\renewcommand\arraystretch{1}
\begin{tabular*}{\columnwidth}{@{\extracolsep{\fill}}lcccccc@{}}
\toprule
 & \multicolumn{3}{c@{}}{Outdoor Scene} & \multicolumn{3}{c@{}}{Indoor scene} \\ 
& PSNR $\uparrow$ & SSIM $\uparrow$ & LPIPS $\downarrow$ & PSNR $\uparrow$ & 
SSIM $\uparrow$ & LIPPS $\downarrow$ \\
\midrule
NeRF~\cite{mildenhall2020nerf} & 21.46 & 0.458 & 0.515 & 26.84 &  0.790 & 0.370 \\
Instant NGP~\cite{muller2022instant} & 22.90 &0.566 & 0.371 & 29.15 & 0.880 & 0.216 \\
MipNeRF360~\cite{barron2022mip360} &  24.47 & 0.691 & 0.283 & \best 31.72 &  0.917 & \tbest 0.180 \\
3DGS~\cite{kerbl3Dgaussians} & \sbest 24.64 & \sbest 0.731 &  \tbest 0.234 & \tbest 30.41 & \tbest 0.920 &  0.189 \\
SuGaR~\cite{guedon2023sugar} & 22.93 & 0.629 &   0.356 & 29.43 & 0.906 & 0.225 \\
2D GS~\cite{Huang2DGS2024} &  24.34 &  0.717 &  0.246 &   30.40 &  0.916 &  0.195 \\
PGSR~\cite{chen2024pgsr} & \tbest 24.45 & \tbest 0.730 & \sbest 0.224 &  \tbest 30.41 & \sbest 0.930 &  \sbest 0.161 \\
\midrule
Ours& \best 24.79& \best 0.751& \best 0.199& \sbest 30.57& \best 0.932& \best 0.153\\
\bottomrule
\end{tabular*}
\label{tab:mipnerf360}
\end{table}

\subsection{Mechanism Analysis of UNG-GS}
\subsubsection{Qualitative Results on the Uncertainty Map} 
\begin{figure*}
    \centering
    \includegraphics[width=\linewidth]{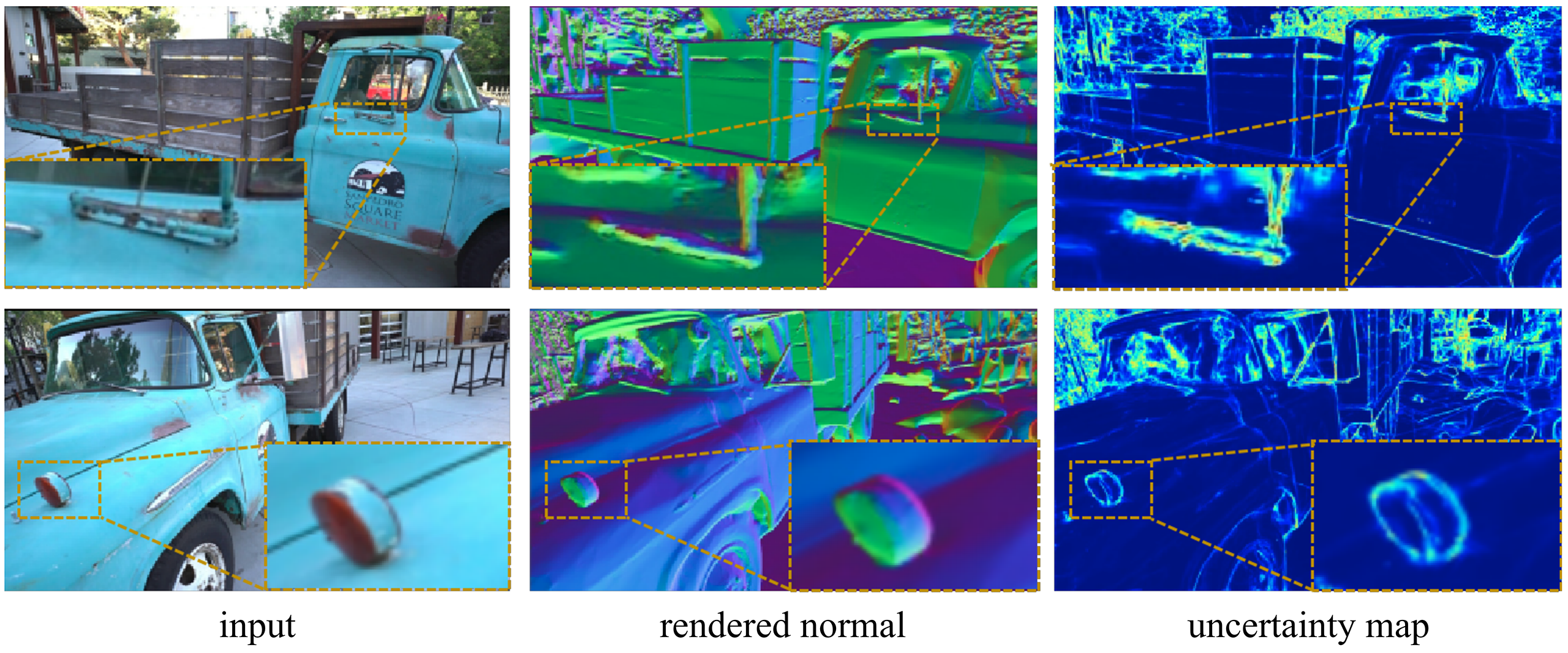}
    \caption{\textbf{Qualitative results of uncertainty map.}}
    \label{fig:re-uncertainty}
\end{figure*}

~\Cref{fig:re-uncertainty} shows that our uncertainty field assigns higher values to geometrically inconsistent regions (e.g., edges and occlusions) while remaining low on flat or low-texture areas. This indicates that SUF captures geometric inconsistency rather than photometric variation.

\subsubsection{Dynamic Learning of the SUF} 
\begin{figure}
    \centering
    \includegraphics[width=\linewidth]{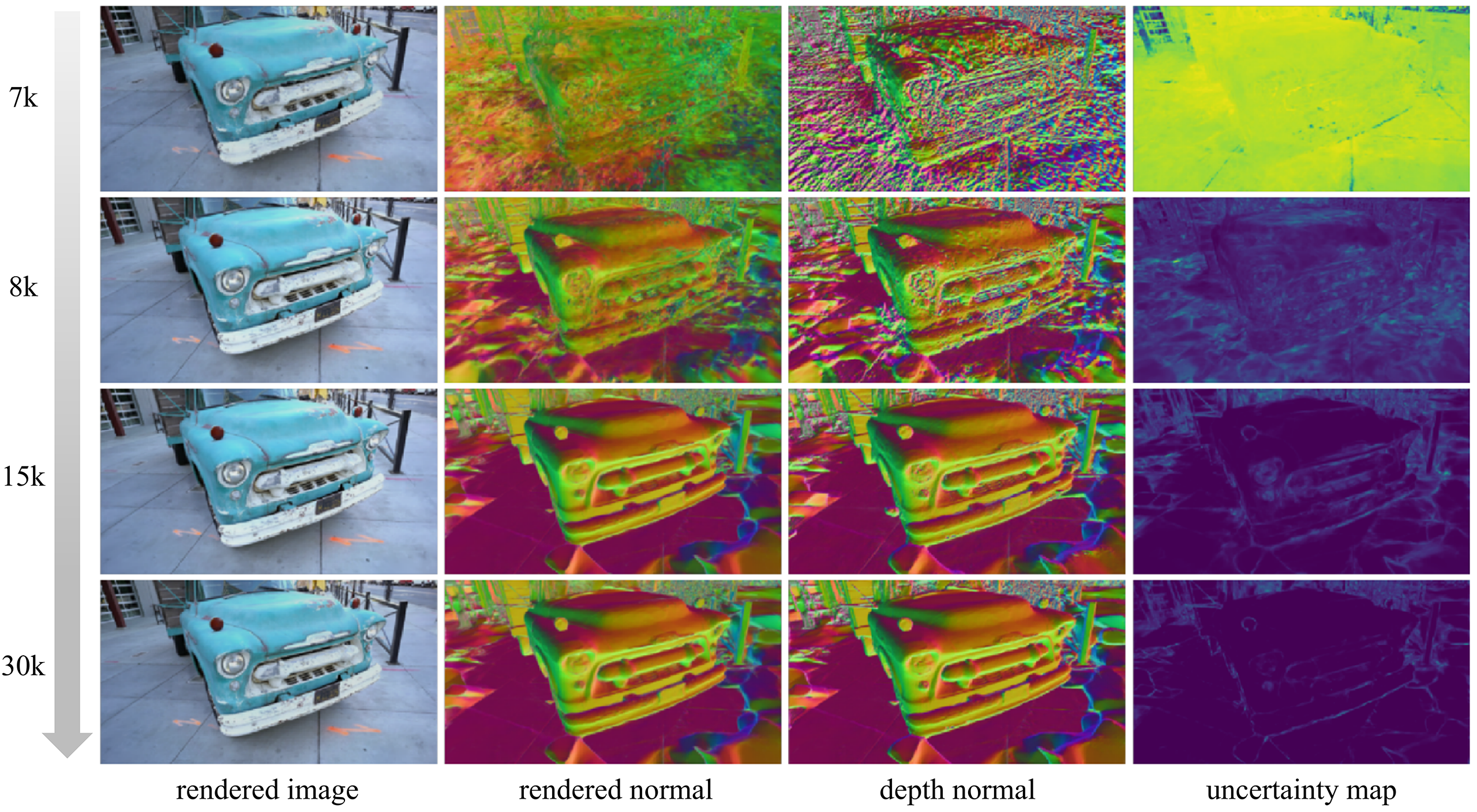}
    \caption{\textbf{Evolution of SUF during training.} As training proceeds, rendering quality improves and uncertainty becomes sparse, localizing to geometrically inconsistent regions.}
    \label{fig:inconsistency_evolution}
\end{figure}

To provide deeper insight into the learning dynamics, \Cref{fig:inconsistency_evolution} visualizes the evolution of key geometric outputs throughout training. In the early stages (e.g., 7k iterations), the geometry is noisy and exhibits widespread conflicts between the rendered normals ($\boldsymbol{\hat{N}}_r$) and depth-derived normals ($\boldsymbol{\hat{N}}_d$), which are reflected by a dense uncertainty map. As optimization proceeds (15k iterations), the geometry becomes more coherent and the uncertainty concentrates on challenging regions such as object boundaries and thin structures. Upon convergence (30k iterations), SUF stabilizes into a precise indicator of residual geometric conflict. This behavior shows that SUF is not a fixed prior, but a dynamically learned field that progressively focuses the optimization on geometrically difficult regions.
\begin{table}
\centering
\caption{\textbf{Components analysis on the TnT Truck.}}
\scriptsize
\setlength{\tabcolsep}{3.5pt}
\renewcommand\arraystretch{1}
\begin{tabular*}{\columnwidth}{@{\extracolsep{\fill}}lccccc@{}}
\toprule
Metric   &  w/o log term&  w/o UGNR&w/o UADR& w/o SUF& full model \\ 
\midrule
F1-Score $\uparrow$ &     0.485 &      0.535&0.530&     0.521&        \textbf{0.560}\\ 
\bottomrule
\end{tabular*}%
\label{tab:ablation}
\end{table}

\subsection{Ablation and Mechanism Validation}
\label{sec:ablation}
\subsubsection{Components Analysis}
\begin{figure}
    \centering
    \includegraphics[width=0.99\linewidth]{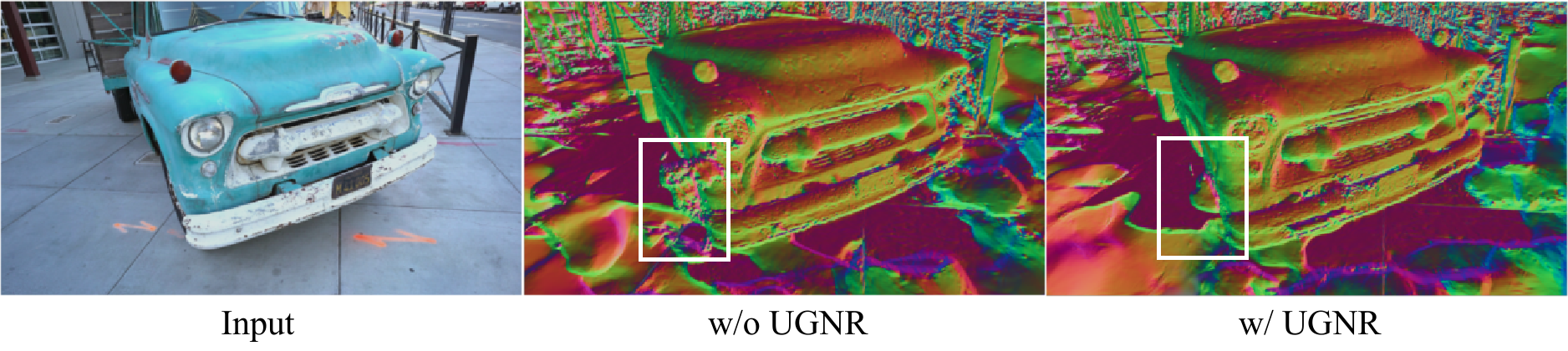}
    \caption{\textbf{Ablation study of UGNR.} We compare the differences in rendered normal images generated during training with and without UGNR after 10000 iterations.}
    \label{fig:ablation-ugnr}
\end{figure}

\Cref{tab:ablation} shows the impact of key components in UNG-GS on the TnT Truck scene. Removing SUF, UADR, or UGNR each degrades performance, with SUF being the most critical because it provides the uncertainty signal used by the downstream geometric modules. We also ablate the $\log(U_i)$ term in the NLL-inspired loss. Without this regularizer, the model predicts excessively large uncertainty scales to minimize the loss, leading to poor normal accuracy, even worse than removing the uncertainty loss entirely. \Cref{fig:ablation-ugnr} further presents a qualitative comparison of rendered normals with and without UGNR, demonstrating its role in improving normal consistency and preserving geometric details.

\subsubsection{Impact of CMF Design}
\begin{figure}
    \centering
    \includegraphics[width=0.8\linewidth]{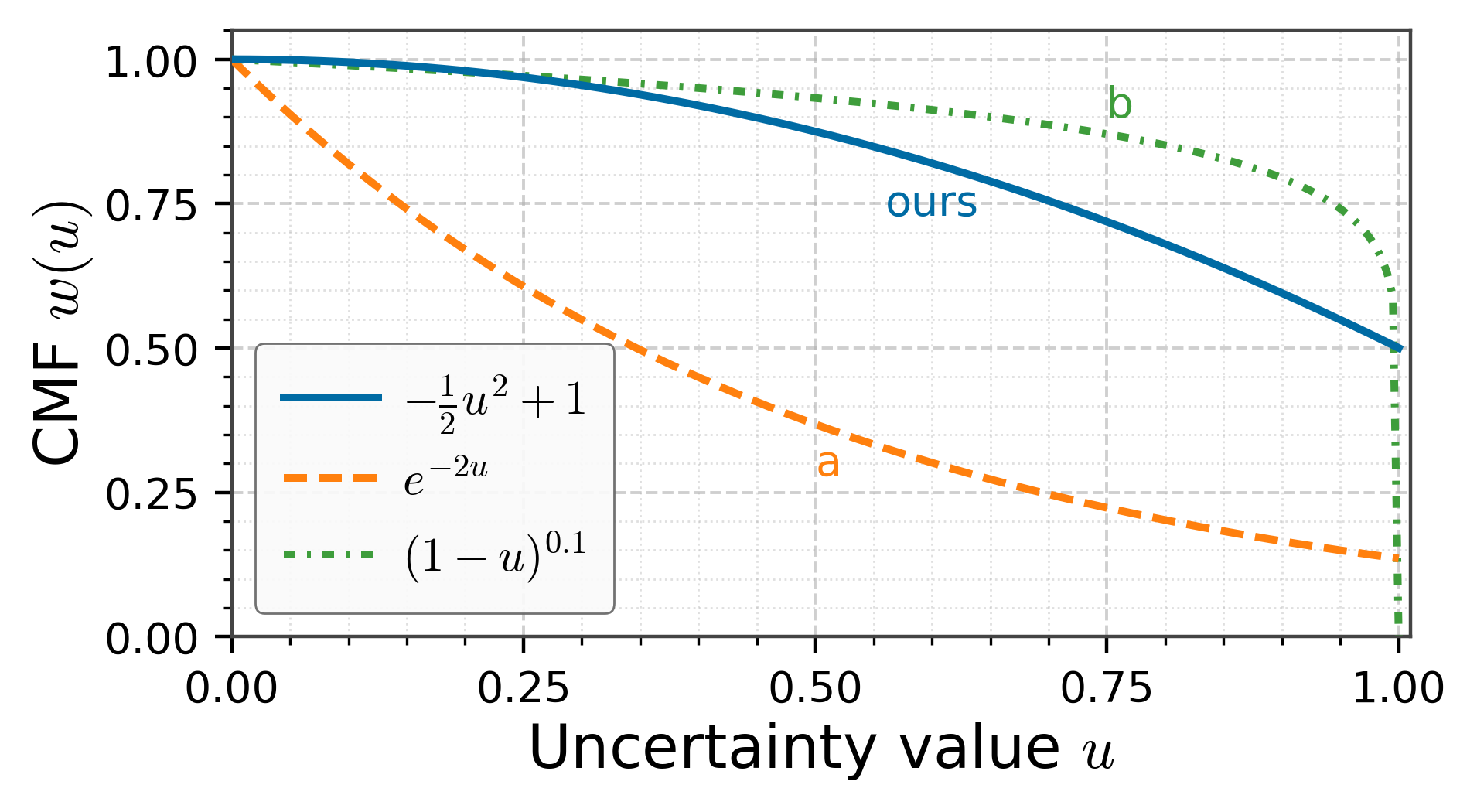}
    \caption{\textbf{Comparison of different CMFs.}}
    \label{fig:weight-functions}
\end{figure}

We compare two alternative CMFs: (a) $w=\exp (-2 u)$ and (b) $w=(1-u)^{0.1}$. As shown in~\Cref{tab:ablation-rendering}, our CMF yields the best performance among the three choices. As illustrated in~\Cref{fig:weight-functions}, this behavior can be attributed to the linear gradient of our CMF. As $u$ increases, the gradient transitions smoothly from 0 to -1, promoting stable optimization and mitigating instability caused by abrupt gradient changes. In contrast, CMF (a) changes more rapidly when $u$ is close to 0, while CMF (b) becomes excessively steep when $u$ approaches 1. Our CMF also retains appropriate weights in low-confidence regions, suppressing noise while preserving useful geometric information.
\begin{table}
\scriptsize
\caption{\textbf{Ablation study of uncertainty-aware depth rendering strategies.} We compare results on the Truck scene.}
\renewcommand\arraystretch{1.1}
\setlength{\tabcolsep}{12.2pt}
\centering
\begin{tabular*}{\columnwidth}{@{\extracolsep{\fill}}lccc@{}}
\toprule
Metric   & (a)~~$\exp (-2 u)$ & (b)~~$(1-u)^{0.1}$ & Ours\\ 
\midrule
F1-Score $\uparrow$ &     0.483&     0.520&     \textbf{0.560}\\ 
\bottomrule
\end{tabular*}%
\label{tab:ablation-rendering}
\end{table}

\begin{table}[t]
\scriptsize
\caption{\textbf{Generalizability of the SUF.} We added SUF to the normal-guided methods PGSR and RaDe-GS to verify the effectiveness and generalization of our proposed framework.}

\renewcommand\arraystretch{1}
\setlength{\tabcolsep}{11pt}
\centering
\begin{tabular*}{\columnwidth}{@{\extracolsep{\fill}}ccccc@{}}
\toprule
\multirow{2}{*}{Metric} & \multicolumn{2}{c}{PGSR~\cite{chen2024pgsr}} & \multicolumn{2}{c}{RaDe-GS~\cite{zhang2024rade}} \\ 
\cmidrule(lr){2-3} \cmidrule(lr){4-5}
 & w/o SUF & w/ SUF & w/o SUF & w/ SUF \\ 
\midrule
F1-Score $\uparrow$ & 0.521 & \textbf{0.543} & 0.329 & \textbf{0.346} \\ 
\bottomrule
\end{tabular*}
\label{tab:ablation-suf}
\end{table}

\subsubsection{Generalizability of the SUF}
\begin{figure}
    \centering
    \includegraphics[width=0.95\linewidth]{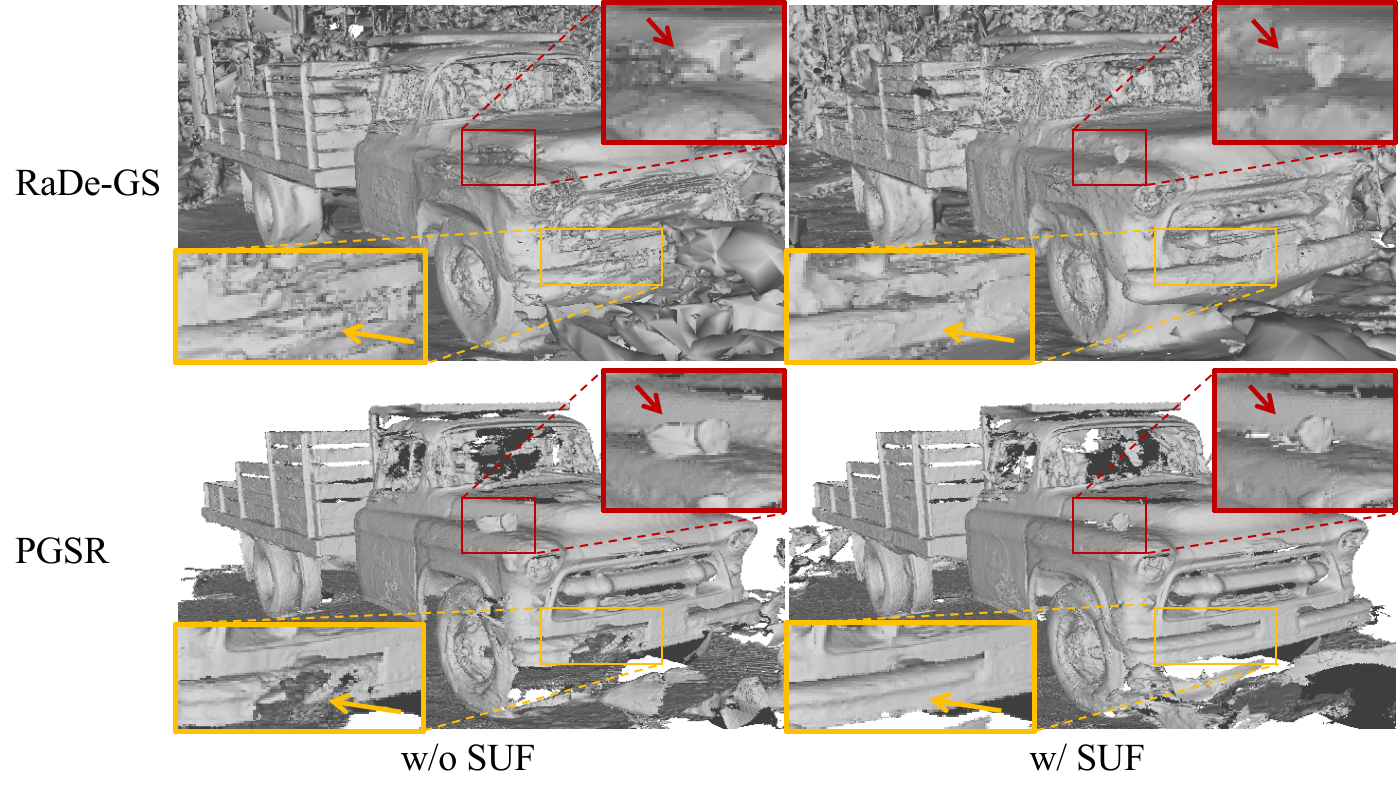}
    \caption{\textbf{Generalizability of the SUF.} We added our proposed SUF to the normal-guided methods PGSR~\cite{chen2024pgsr} and RaDe-GS~\cite{zhang2024rade} for verification and found that SUF can better reconstruct edge features and is not prone to losing details.}
    \label{fig:ablation-suf}
\end{figure}

To assess the generalization ability of SUF, we integrate it into two existing normal-guided surface reconstruction methods, PGSR~\cite{chen2024pgsr} and RaDe-GS~\cite{zhang2024rade}. We modify the publicly available CUDA implementations of these methods, incorporate SUF, and retrain the models. As shown in~\Cref{tab:ablation-suf} and visualized in~\Cref{fig:ablation-suf}, SUF consistently improves reconstruction accuracy for both PGSR and RaDe-GS. These results suggest that explicitly modeling geometric uncertainty is not specific to UNG-GS, but can benefit diverse normal-guided reconstruction pipelines.

\begin{figure}
    \centering
    \includegraphics[width=\linewidth]{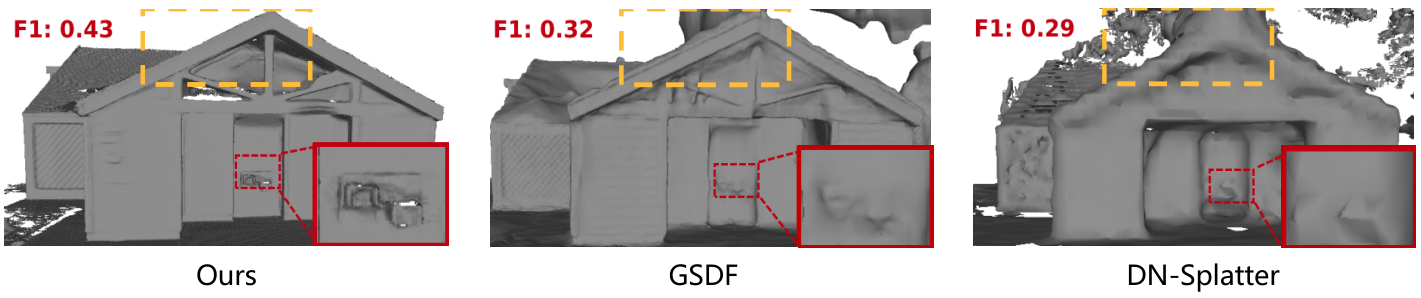}
    \caption{\textbf{Qualitative Comparisons with GSDF and DN-Splatter.}}
    \label{fig:re-more-results}
\end{figure}

\subsection{Extended Analyses}
\label{sec:extended_experiments}

To further assess the scope of UNG-GS beyond the main sparse-sequence reconstruction results, we conduct extended analyses on prior-based comparisons and dense-sequence reconstruction.

\subsubsection{Comparison with Prior-Based Methods}
We further compare UNG-GS with methods that rely on external geometric priors, including priors derived from foundation models. As shown in~\Cref{fig:re-more-results}, our method recovers sharper geometric details than DN-Splatter, supporting our claim that accurate surface reconstruction can be achieved without relying on such priors. We also explore incorporating prior normals as additional supervision. As shown in~\Cref{tab:ablation-priors}, adding prior normals to UNG-GS further improves performance. However, our baseline without prior supervision still outperforms the setting that uses prior normals without SUF. This indicates that the main gain comes from uncertainty modeling, while prior normals provide complementary benefits.
\begin{table}[t]
\scriptsize
\centering
\caption{\textbf{Effect of incorporating prior normals on the Truck scene in TnT\@.} We explore prior normals as additional supervision with or without SUF. $\Delta$ denotes the change relative to Ours (without prior, with SUF).}\label{tab:ablation-priors}
\renewcommand{\arraystretch}{1.05}
\setlength{\tabcolsep}{6pt}
\begin{tabular*}{\columnwidth}{@{\extracolsep{\fill}}lcc@{}}
\toprule
Setting & F1-score $\uparrow$ & $\Delta$ \\
\midrule
With prior, without SUF & 0.501 & -0.059 \\
Ours (without prior, with SUF) & 0.560 & \textemdash \\
Ours (with prior, with SUF) & \textbf{0.566} & +0.006 \\
\bottomrule
\end{tabular*}
\end{table}

\begin{table}[t]
\centering
\caption{\textbf{Quantitative comparison on the standard dense-sequence settings of DTU and TnT.} For brevity, we report only dataset-level means. Chamfer Distance (CD) [mm$\downarrow$] is reported on DTU and F1-score ($\uparrow$) on TnT. Baseline results are taken from~\cite{chen2024pgsr} except for RaDe-GS~\cite{zhang2024rade}, whose results are from the original paper.}
\scriptsize
\setlength{\tabcolsep}{4pt}
\renewcommand{\arraystretch}{1}
\begin{tabular*}{\columnwidth}{@{\extracolsep{\fill}}lcc@{}}
\toprule
Method & DTU CD [mm]$\downarrow$ & TnT F1$\uparrow$ \\
\midrule
NeuS~\cite{wang2021neus} & 0.84 & 0.38 \\
Neuralangelo~\cite{li2023neuralangelo} & \tbest0.61 & \tbest0.50 \\
\midrule
3DGS~\cite{kerbl3Dgaussians} & 1.96 & 0.09 \\
SuGaR~\cite{guedon2023sugar} & 1.33 & 0.19 \\
2DGS~\cite{Huang2DGS2024} & 0.80 & 0.32 \\
GOF~\cite{Yu2024GOF} & 0.74 & 0.46 \\
RaDe-GS~\cite{zhang2024rade} & 0.68 & 0.37 \\
PGSR~\cite{chen2024pgsr} & \sbest0.52 & \best0.52 \\
\midrule
Ours & \best0.51 & \sbest0.51 \\
\bottomrule
\end{tabular*}
\label{tab:dense_origin}
\end{table}

\subsubsection{Dense-Sequence Robustness Evaluation}
We further evaluate UNG-GS on the standard dense-sequence settings of the DTU and TnT datasets. As shown in~\Cref{tab:dense_origin}, UNG-GS remains competitive under dense-view inputs, achieving the lowest mean CD of 0.51 mm on DTU and a comparable mean F1-score of 0.51 on TnT. For brevity, we report only dataset-level means, since this experiment is intended to test robustness beyond the primary sparse-sequence setting. These results suggest that UNG-GS does not rely on sparse-only assumptions, even though its main advantage appears in sparse-sequence reconstruction.

The performance gain under dense image sequences is smaller than that in sparse settings. We attribute this to the substantially larger overlap across views in dense sequences. On the one hand, stronger overlap already improves geometric consistency and provides stronger supervision for reconstruction. On the other hand, it reduces overall geometric uncertainty, making explicit uncertainty modeling less critical than in sparse-sequence scenarios. As a result, UNG-GS remains competitive under dense-view settings, but its advantage is more pronounced when the input image sequence is sparse.

\begin{figure}[t]
    \centering
    \includegraphics[width=0.98\linewidth]{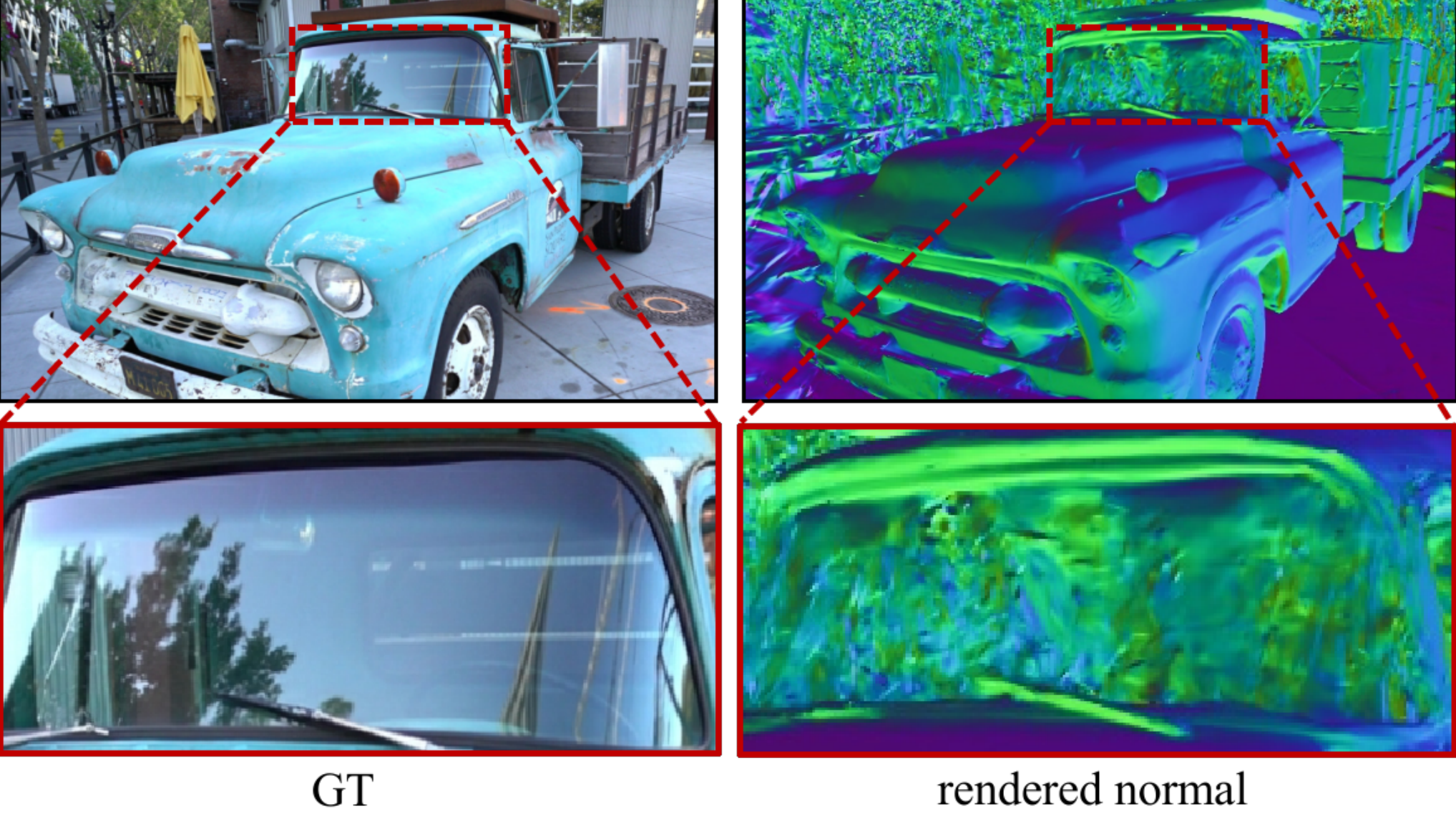}
    \caption{\textbf{Glass mirror reflection reconstruction failed.}}
    \label{fig:limitation}
\end{figure}

\section{Discussion \& Limitations}
UNG-GS achieves significant improvements in sparse-view reconstruction.
Additional analyses further suggest that the proposed SUF and related strategies are not restricted to sparse-only settings, although their benefits are most evident when geometric uncertainty is high.
However, as shown in~\Cref{fig:limitation}, UNG-GS may still face challenges with highly reflective and transparent surfaces, such as those found in mirrors and glass. Future work could explore incorporating additional priors or specialized rendering methods to better handle these complex materials.

\section{Conclusion}
\label{sec:conclusion}
This paper presents UNG-GS, a novel framework for high-fidelity surface reconstruction and novel view synthesis from sparse image sequences that, in its default setting, does not rely on priors. By explicitly modeling geometric uncertainty through a normal-guided Spatial Uncertainty Field, UNG-GS tackles the limitations of existing 3DGS methods, which suffer from degraded reconstruction quality due to the lack of uncertainty quantification.
Our key innovation lies in embedding probabilistic modeling into the 3DGS framework, enabling adaptive error tolerance during optimization. The proposed uncertainty-aware rendering strategy dynamically suppresses noise in uncertain regions while preserving details in high-confidence areas, and the uncertainty-guided adaptive normal refinement robustly aligns surface normals by adjusting neighborhood gradient weights.
Extensive experiments on public datasets demonstrate that UNG-GS consistently improves sparse-sequence reconstruction.

\bibliographystyle{IEEEtran}
\bibliography{refs}

\end{document}